\documentclass{article}

\usepackage{comment}
\usepackage{amsmath}
\usepackage{arxiv}
\usepackage{algpseudocode}
\usepackage[ruled,vlined]{algorithm2e}
\usepackage{arydshln}
\usepackage{float}

\usepackage[utf8]{inputenc} % allow utf-8 input
\usepackage[T1]{fontenc}    % use 8-bit T1 fonts
\usepackage{hyperref}       % hyperlinks
\usepackage{url}            % simple URL typesetting
\usepackage{booktabs}       % professional-quality tables
\usepackage{amsfonts}       % blackboard math symbols
\usepackage{nicefrac}       % compact symbols for 1/2, etc.
\usepackage{microtype}      % microtypography
\usepackage{lipsum}
\usepackage{graphicx}
\graphicspath{ {./images/} }

\title{EXACFS - A CIL Method to Mitigate Catastrophic Forgetting}
\author{
S Balasubramanian \\
Department Of Mathematics and Computer Science\\
Sri Sathya Sai Institute of Higher Learning\\
Puttaparthi, Andhra Pradesh, India\\
\texttt{sbalasubramanian@sssihl.edu.in} \\\AND
M Sai Subramaniam \\
Department Of Mathematics and Computer Science\\
Sri Sathya Sai Institute of Higher Learning\\
Puttaparthi, Andhra Pradesh, India\\
\texttt{saimurali2001@gmail.com} \\\And
Sai Sriram Talasu \\
Department Of Mathematics and Computer Science\\
Sri Sathya Sai Institute of Higher Learning\\
Puttaparthi, Andhra Pradesh, India\\
\texttt{saisriramtalasu206@gmail.com} \\\And
Yedu Krishna P \\
Department Of Mathematics and Computer Science\\
Sri Sathya Sai Institute of Higher Learning\\
Puttaparthi, Andhra Pradesh, India\\
\texttt{p.yedukrishna007@gmail.com} \\\And
Manepalli Pranav Phanindra Sai \\
Department Of Mathematics and Computer Science\\
Sri Sathya Sai Institute of Higher Learning\\
Puttaparthi, Andhra Pradesh, India\\
\texttt{pranavnag257@gmail.com} \\\And
Ravi Mukkamala \\
School of Coumputing and Information\\
Old Dominion University\\
\texttt{rmukkama@odu.edu} \\\And Darshan Gera \\
Department Of Mathematics and Computer Science\\
Sri Sathya Sai Institute of Higher Learning\\
Puttaparthi, Andhra Pradesh, India\\
\texttt{darshangera@sssihl.edu.in} 
}

\begin{document}
\maketitle
\begin{abstract}
Deep neural networks (DNNs) excel at learning from static datasets but struggle with continual learning, where data arrives sequentially. Catastrophic forgetting, the phenomenon of forgetting previously learned knowledge, is a primary challenge. This paper introduces EXponentially Averaged Class-wise Feature Significance (EXACFS) to mitigate this issue in the class incremental learning (CIL) setting. By estimating the significance of model features for each learned class using loss gradients, gradually aging the significance through the incremental tasks and preserving the significant features through a distillation loss, EXACFS effectively balances remembering old knowledge (stability) and learning new knowledge (plasticity). Extensive experiments on CIFAR-100 and ImageNet-100 demonstrate EXACFS's superior performance in preserving stability while acquiring plasticity. 
\end{abstract}

% keywords can be removed

\section{Introduction}
Class Incremental Learning (CIL) is vital for real-world applications where data arrives sequentially, necessitating continuous model adaptation with extremely limited or no access to previous data. Examples include progressively learning new image categories in image recognition, expanding vocabulary and grasping novel language patterns in natural language processing, and adapting to new traffic scenarios and road conditions in autonomous vehicles.
\newline
\begin{figure}
    \centering
    \includegraphics[scale=0.5]{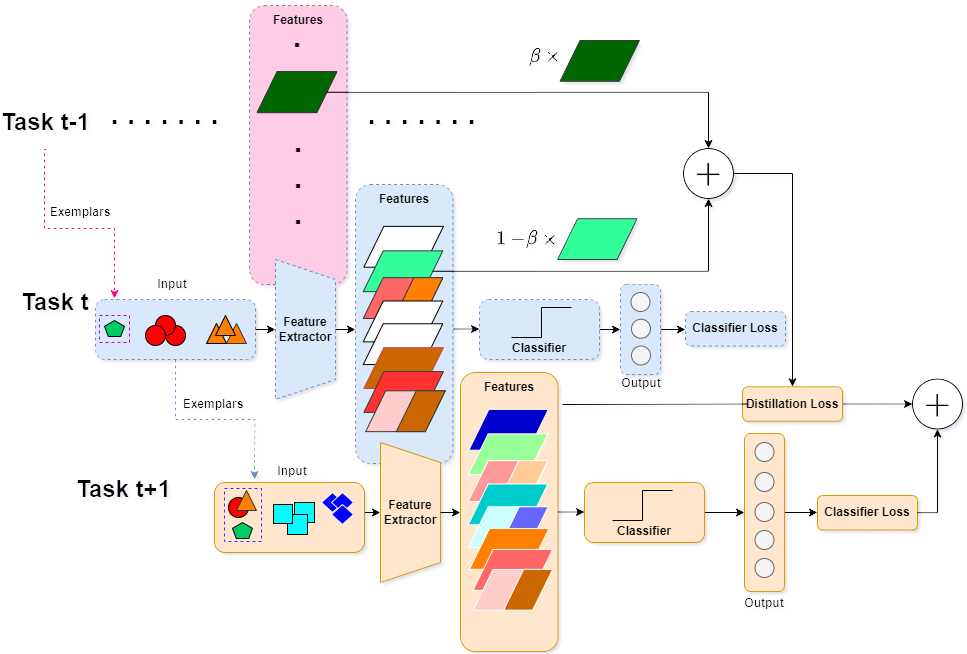}
    \caption{\textbf{Schematic of the EXACFS Method. 
    At incremental task $t+1$, input comprises of samples from new classes and a few exemplars from classes of earlier tasks. Class labels are shown by colour coding. A shared but incrementally updated feature extractor extracts features. Features influencing class(es) decision are accordingly colour-coded, with the intensity of colour conveying the level of significance. Colourless features do not influence any input classes. A feature may influence more than one class with different degrees. During training of task $t+1$, along with classifier loss, a distillation loss constraining features of exemplars to be similar to its earlier representation from task $t$ is aggregated to promote stability. Depending on the class the exemplar belongs to, the feature similarity across tasks $t$ and $t+1$ is accordingly weighted by an exponentially averaged feature significance. Refer to section \ref{sec:EXACFS} for more details.}}
    \label{fig:teaserdiag}
    \vspace{-13pt}
\end{figure}
A primary challenge in CIL is catastrophic forgetting, where the model's performance on previously learned tasks deteriorates significantly as it acquires new knowledge.  This phenomenon arises from the inherent instability of neural networks, as updates to the model's parameters to accommodate new classes can inadvertently overwrite information crucial for recognizing old ones. Overcoming catastrophic forgetting is essential for developing practical CIL systems that can continuously learn and adapt to changing environments without compromising past performance.
\newline
\newline
Various approaches have been proposed to mitigate catastrophic forgetting in class incremental learning (CIL). Regularization methods \cite{EWC, SI, MAS} introduce constraints during training to limit changes to critical model parameters, thereby preserving previously learned knowledge. Replay methods \cite{ICARL, GEM, DGR} address forgetting by storing and reusing samples from old classes, called as exemplars, either through actual data (experience replay) or synthetic data (generative replay). Architectural methods \cite{PN, AANET, DER} modify the network’s structure to support new classes, such as by adding new pathways while maintaining existing ones. Distillation-based approaches in CIL focus on preserving knowledge of previous models. These methods involve transferring knowledge from an old model to a new one, ensuring that the new model mimics the output distributions of the previous model \cite{SSIL, ICARL, LWF}, which helps maintain task performance over time. Additionally, distillation techniques can be adapted to preserve feature representations  \cite{LUCIR, PODNET, AFC} by constraining the new model to maintain similarity in feature activations between old and new tasks. Meta-learning techniques \cite{ML} aim to enable models to adapt swiftly to new classes with minimal forgetting, leveraging prior experiences. While many of the methods from the literature are hybrid in nature, they can generally be categorized into one of the aforementioned approaches. Collectively, these methods strive to balance the retention of old knowledge (called stability) with the acquisition of new information (called plasticity), effectively addressing the challenge of catastrophic forgetting in CIL.
\newline
In this work, we introduce EXponentially Averaged Class-wise Feature Significance (EXACFS), a novel distillation based approach designed to mitigate catastrophic forgetting in CIL. By assigning class-specific significance to feature maps and enforcing their preservation during incremental training, EXACFS effectively balances the stability with the plasticity. The significance of features is exponentially updated, ensuring that past values are gradually aged while preserving their importance throughout the learning process. Although state-of-the-art methods (SOTA) \cite{LUCIR, PODNET} focus on maintaining feature similarity across increments, EXACFS sets itself apart by variably attending to features based on their significance and estimating feature significance on a class-wise basis, rather than uniformly across all the classes. Overall, our contributions are as follows.
\begin{enumerate}
\item We introduce a class-wise feature significance approach for feature preservation, effectively retaining previously learned knowledge while integrating new information in CIL settings.
\item We implement an exponential averaging of feature significance, ensuring that past values gradually age to enhance stability.
\item We propose a novel feature distillation loss objective tailored for incremental training.
\item We demonstrate the effectiveness and superiority of our method through extensive experiments and ablation studies on benchmark datasets such as CIFAR-100 and ImageNet-100, showcasing its advantages over SOTA methods.
\end{enumerate}

% \vspace{.3cm}
\noindent
The rest of the paper is organized as follows. Section 2 reviews related work. In Section 3, we provide a detailed description of our proposed method, EXACFS. Section 4 covers the datasets, experimental setup, results, and various ablation studies. Finally, Section 5 presents the conclusions.

\section{Related Work}
\par Incremental learning is an active area of research in the broader field of deep learning. Recently, numerous methods have been proposed towards making the neural networks learn incrementally. These methods are broadly classified into the  following categories: regularization, replay, architecture, distillation, and meta-learning. We review important methods from these categories.
\newline
\newline
\textbf{Regularization-based:}
Regularization approaches aim to either encourage or discourage updates to specific parameters of the model based on their significance. For example, Elastic Weight Consolidation (EWC) \cite{EWC} assesses each parameter's rigidity or flexibility using the Fisher information matrix, utilizing its relevance to past tasks to guide model updates. Similarly, Synaptic Intelligence (SI) \cite{SI} assigns importance to each parameter by calculating the path integral along the optimization path. Memory Aware Synapses (MAS) \cite{MAS} computes the importance of parameters in an unsupervised manner by assessing the sensitivity of the learned function to changes in parameters. It helps to protect important weights during learning. A Bayesian framework relying on Laplace approximations to estimate parameter importance is presented in \cite{BLA}. Research has shown that parameter regularization methods, while useful, often yield lower accuracy compared to other techniques \cite{AFC16, AFC46}. These methods may not be as effective in maintaining the integrity of network outputs. Consequently, they might serve as a less reliable means for preserving learned knowledge \cite{AFC8}.
\newline
\newline
\textbf{Replay-based:}
Experience Replay (ER) methods \cite{ER1, ER2} store and replay samples from old classes during the training of new classes, improving memory efficiency and model performance through intelligent sampling strategies. Generative Replay approaches, such as DGR \cite{DGR}, utilize generative models to synthesize samples of previously learned classes, thus reducing memory overhead and mitigating privacy concerns. iCaRL \cite{ICARL} combines replay with knowledge distillation, using a herding algorithm to select representative exemplars. BiC \cite{BIC} introduces a two-stage training process to address class imbalance, incorporating a bias correction layer to adjust model logits. GEM \cite{GEM} and its extension A-GEM \cite{AGEM}, integrate replay with gradient constraints to ensure non-interference with past task gradients. Gradient Sample Selection \cite{GSS} optimizes sample selection based on gradient diversity, enhancing the replay buffer's representativeness. Replay based regularization methods like ER-ACE \cite{ERACE} combine replay with parameter regularization to balance knowledge retention and new learning. Other methods like \cite{Mnemonics} parametrize the exemplars for replay and make them optimiziable using a bi-level optimization in an end-to-end manner. We will employ replay techniques as recommended in iCaRL \cite{ICARL} in this work. 
\newline
\newline
\textbf{Architecture-based:}
Architecture-based methods for CIL focus on evolving network structures to incorporate new tasks while preserving previously acquired knowledge. Progressive Networks \cite{PN} retain a pool of pretrained models and learn lateral connections to leverage their features for new tasks. This approach integrates prior knowledge across the feature hierarchy, allowing the model to combine old and new information effectively. By adding new capacity, these networks can reuse previous computations while adapting to new ones. AANet \cite{AANET} utilizes plasticity and stability blocks to strike a balance between learning new information and maintaining old knowledge. Plasticity blocks facilitate adaptation to new tasks, while stability blocks ensure the retention of previously learned features. DER (Dynamically Expandable Representation) \cite{DER} enhances the network's capacity by introducing a new feature extractor for each incremental task and keeping the feature extractors of old tasks frozen. This approach expands the network's representational ability for new classes while preserving learned representations. To optimize performance and reduce redundancy, DER employs a differentiable channel-level mask-based pruning strategy. Partial DER \cite{SPD} refines this approach by applying DER exclusively in the last residual block of the base network, thus controlling architectural complexity while maintaining incremental learning efficiency. 
\newline
\newline
\textbf{Distillation-based:}
In CIL, distillation based methods are designed to safeguard both the output distributions and feature representations from previous models. These techniques work by transferring knowledge from a prior model to a newer one, ensuring that the updated model replicates the output distributions of its predecessor, thus maintaining consistent task performance over time. Furthermore, these distillation strategies can also be tailored to preserve feature representations, requiring the new model to maintain similarity in feature activations between previously learned and new tasks. iCaRL \cite{ICARL} constrains old class distributions between old and new model across new class samples and old exemplars to be similar. LUCIR \cite{LUCIR} forces feature representations of both exemplars and new class samples to be similar between old and new model, fixing the old class embeddings. SSIL \cite{SSIL} first separates the class distribution estimation between old and new classes, and subsequently incorporates task based knowledge distillation to mitigate catastrophic forgetting. PODNet \cite{PODNET} proposes a feature distillation strategy based on step-wise spatial aggregation of convolutional features, first along width and then along height dimensions. For the dense layer, the feature distillation is similar to the one proposed in \cite{LUCIR}. In AFC \cite{AFC}, feature similarity between old and new models is imposed by restricting the loss with respect to features to be similar between both the models. This is achieved by minimizing an upper bound of difference in the loss between both the models, arrived through a first order approximation. In both AFC and PODNet, distillation is applicable through all the layers of the model. A novel distillation approach that first constructs low-dimensional manifolds for previous and current responses and subsequently minimizes the dissimilarity between the responses along the geodesic connecting the manifolds is elucidated in \cite{GeoDL}.
\newline
\newline
Our proposed EXACFS is also a feature distillation approach applicable through all layers of the model. The details of our method is presented in section \ref{sec:EXACFS}.
\newline
\newline
\textbf{Meta-Learning-based:}
In \cite{ML}, the authors present continual learning as a meta-learning challenge that seeks to balance catastrophic forgetting with the ability to acquire knowledge from new tasks. They introduce the meta-experience replay (MER) algorithm, which integrates the Reptile meta-learner with experience replay to address this challenge. The connection to Reptile simplifies the approach, achieving effective continual learning without the need for identification of tasks, making the algorithm broadly applicable.

\section{The Proposed Method EXACFS}
\label{sec:EXACFS}
\subsection{CIL Setting}
There are $N+1$ tasks, the first being the initial task and the rest arriving incrementally, with each task $t$ comprising a set of $n^t$samples belonging to one of $c^t$ different classes. Usually, $c^t$ is large for the initial task and uniform for rest of the tasks. The classes themselves are non-overlapping across all the tasks. Formally, let $D^t = \{(x_i, y_i)\}_{i=1 \sim n^t}$ be the data for task $t$ where $t \in \{0, 1, ..., N\}$ and $y_i \in \{r^{t-1}+1, ..., r^t$ $\mid$ $r^t - r^{t-1} = c^t\}$ where $r^{-1} = 0$. Let $D = \bigcup_{t=0}^{N}D^t$ and $n=\sum_{t=0}^{N}n^t$. From task $1$ onwards, we assume availability of a few exemplars from the classes of all the previous tasks. Specifically, let $E^{t-1}$ be the set of exemplars from task $t-1$ where $t \in \{1, 2, ..., N\}$. $E^{t-1}$ has $e^i$ exemplars per class $i$ where $i$ ranges through all the class labels in task $t-1$, i.e., $i \in \{r^{t-2}+1, \ldots, r^{t-1} \,\,|\,\, r^{t-1} - r^{t-2} = c^{t-1}\}$. Let $E^{0 \sim t-1} = \bigcup_{i=0}^{t-1}{E^i}$ where $t \in \{1, 2, ..., N\}$ be the set of exemplars available to task $t$. It is assumed that $e^i \ll n^{t}$. At the end of training each task $t$, we evaluate the trained model on the test sets of current task and all the previous tasks. We repeat this training and testing procedure through the $N+1$ tasks.
\newline
\newline
With regard to architecture/model denoted by $M$, we assume a fixed convolutional network $F$ across the tasks, appended with a dense linear layer $P$ and a growing classifier layer $g$ that adds $c^t$ nodes at task $t$ to the already existing nodes representing classes from the previous tasks. $F$ is assumed to have $L-1$ layers whose outputs are the $L-1$ features $f_1$, $f_2$, ..., $f_{L-1}$, respectively. The features from $P$ are denoted by $f_L$. At task $t$, $M$, $F$, $P$ and $g$ are denoted by $M^t$, $F^t$, $P^t$ and $g^t$, respectively. Correspondingly, the feature $f_i$ is denoted as $f_i^t$ at task $t$. At task $t$, the parameters of $F$, $P$ and part of $g$ corresponding to the classes from previous tasks are initialized from the trained model at task $t-1$. The rest of the parameters of $g$ are initialized randomly.

\subsection{The Feature Significance Estimation} \label{feat_sig_estimation}
Training the model at task $t$ by initializing it from the model trained at task $t-1$ and using a standard classifier loss like cross-entropy corresponds to fine-tuning. Mere fine-tuning, in the presence of severe data imbalance—characterized by a few exemplars of old classes versus a large number of samples of new classes—leads to classification score bias \cite{SSIL}. This bias pushes the model to favor new classes, resulting in catastrophic forgetting.
\newline
\newline
To mitigate this, we ensure the preservation of feature representations of exemplars during incremental task training. Similar approaches, such as those in \cite{LUCIR, PODNET}, also aim to preserve feature representations. However, these methods attempt to uniformly preserve all feature representations, which impedes the model's plasticity to learn new regularities introduced by new class samples. While uniform preservation promotes stability, it hinders plasticity.
\newline
\newline
Our approach differs from \cite{LUCIR, PODNET} by estimating the significance of feature representations. We enhance stability by preserving significant features and promote plasticity by allowing other features to capture novel structures in new classes. Additionally, unlike \cite{LUCIR}, our method applies to feature representations at all layers, not just the penultimate dense layer. Next, we will explain the estimation of feature significance.
\newline
\newline
A feature is significant with respect to a class if a small perturbation to it causes an arbitrary sample of the class to be misclassified. This sensitivity of the feature is captured by the gradient of the loss with respect to the feature. Specifically, the magnitude of the gradient can quantify the sensitivity or significance of the feature. Given this understanding, we formally define the feature significance as follows.
\newline
\newline
Let $\mathcal{X}^c = \{(x_i^c, y_i^c)\}$ be the data samples belonging to class $c$ at task t i.e. $y_i^c = c$. $c$ could be one of the classes in the previous tasks $0$ to $t-1$ or from task $t$ itself. In the former case, $\mathcal{X}^c$ comes from $E^{0 \sim t-1}$; in the latter case, $\mathcal{X}^c$ comes from $D^t$. Note that the prediction $\hat{y}_i^c$ for the data $x_i^c$ is given by:

\begin{equation*}
    \hat{y}_i^c = M^t(x_i^c)
                = g^t(P^t(F^t(x_i^c))) 
                = g^t(f_L^t(f_{L-1}(....(f_1^t(x_i^c)))) \\
\end{equation*}  

Let ${CL}^t$ be the classification loss function at task $t$. Then
\begin{equation*}
    {CL}^t = \frac{1}{|D^t|}\sum_{k=1}^{|D^t|}l_{k}^t \\
\end{equation*} 

\noindent where $l_{k}^t$ is the $k^{th}$ sample classification loss at task $t$ and $|D^t|$ is the number of samples in task $t$.
\newline
\newline
Let $\nabla_{k}^c f_{j}^t$ denote the gradient of $l^t_k$ with respect to feature $f_j^t$ for an arbitrary sample indexed by $k$ belonging to class $c$. %A convolutional feature map $f_j^t$ is of size $d_j \times h_j \times w_j$ where $d_j$ is the number of channels and $h_j$, $w_j$ refer to the grid size. Correspondingly $\nabla f_j^t$ will also be of size $d_j \times h_j \times w_j$. The feature map $f_L^t$ is of size $d_L$. Correspondingly $\nabla f_L^t$ will also be of size $d_L$. Now, we define the per-class gradient of ${CL}^t$ with respect to feature map $f_j^t$ for class $c$ as: 
Then the feature significance of feature $f_j^t$ with respect to class $c$ at task $t$ is defined as:
\begin{equation}
        \label{eq: feature_significance}
        S_j^{t,c} = \frac{1}{|\mathcal{X}^c|} \sum_{k:\, x_k \in \mathcal{X}^c} {{\left(\nabla_{k}^{c} f_j^t\right)}^2} 
\end{equation}
\noindent where $|\mathcal{X}^c|$ is the size of $\mathcal{X}^c$ and the meaning of equation \eqref{eq: feature_significance} becomes more lucid depending on the type of feature viz. convolutional or dense, as elucidated in the following two paragraphs.
\newline
\newline
A feature $f_j^t$ from a convolutional layer ($j \in \{ 1, \ldots L-1\}$) has size $d_j \times h_j \times w_j$ for an arbitrary sample indexed by $k$, where $d_j$ is the number of channels and $h_j$, $w_j$ refer to the grid size. A specific channel, say channel $q$, denoted by ${f_j}_q$ has size $h_j \times w_j$. Corresponding to the size of $f_j^t$, $\nabla_{k}^{c} f_{j}^t$ also has size $d_j \times h_j \times w_j$. In this case, we will compute $d_j$ significances. To accomplish this, we first collapse the grid dimensions by averaging $\nabla_{k}^{c} f_{j}^t$ across $h_j$ and $w_j$. This will result in a $d_j$-dimensional vector. For each component in this vector, we compute the significance using equation \eqref{eq: feature_significance}. Specifically, for component $q$, the significance is denoted by $S_{j_q}^{t,c}$.
\newline
\newline
Feature $f_L^t$ from layer $P$ for an arbitrary sample indexed by $k$ will be a vector of size $d_L$. Correspondingly $\nabla_{k}^{c} f_{L}^t$ will also be a vector of size $d_L$. Again, in this case, we will compute $d_L$ significances by using equation \eqref{eq: feature_significance} on each component of this vector. Specifically, for component $q$, the significance is denoted by $S_{L_q}^{t,c}$.

\subsection{Incorporating History}
As old classes percolate through the incremental tasks with only minimal representation by a set of exemplars, it is critical for stability that the class-wise significance of the feature computed using the entire training set of its corresponding task earlier is preserved through the increments. However, retaining the old significance unchanged would hinder the plasticity of the model. We propose a solution to this stability-plasticity trade-off through exponential averaging, where the old feature significance is gradually aged. Formally, let $\beta \in [0, 1]$. Then, 
\begin{equation}
    \widetilde{S}_j^{t,c} = \beta \cdot \widetilde{S}_j^{t-1, c} \, + \, (1 - \beta) \cdot S_j^{t, c}
\end{equation}
\noindent where $t \in \{1, \ldots, N-1\}$ and $\widetilde{S}_j^{0, c} = S_j^{0, c}$.
\newline
\newline
For convolutional and dense features, the corresponding component-wise notation with respect to exponentially averaged significances is similar to what is described in section \ref{feat_sig_estimation}, respectively. Algorithm \ref{alg : EXACFS} presents the steps to compute feature significances.

\subsection{Training EXACFS}
To train the EXACFS model, we aggregate the classification loss with a novel feature distillation proposed as follows. The distillation for feature $j$ during training of task $t$ is defined as:
\begin{equation}
    \label{eq:distillation_loss_term}
    {DL}^t_j = \frac{1}{|E^{0 \sim t-1}|} \sum_{k:\,\, x_k \in E^{0 \sim t-1}}{ \left \langle
                                    \sum_{c=1}^{r^{t-1}}{\widetilde{S}^{t-1,c} \cdot \textbf{1}(y_k = c)} , \,\,
                                    \Delta{f_j^t}        
                                    \right \rangle
                                 }
\end{equation}
where
\begin{align*}
    & \widetilde{S}^{t,c} = \left( \widetilde{S}_{j_1}^{t,c}, \, \ldots \, , \widetilde{S}_{j_{d_j}}^{t,c} \right)^T \\
    & \Delta{f_j^t} = \left( || f_{j_1}^t - f_{j_1}^{t-1} ||^2_F, \, \ldots \, , || f_{j_{d_j}}^t - f_{j_{d_j}}^{t-1} ||^2_F \right)^T ,
\end{align*}

\textbf{$\textbf{1}\left( \cdot \right)$} denotes the indicator function and $||\cdot||_F$ is the matrix Frobenius norm; $||\cdot||_F$ degenerates to absolute value $|\cdot|$ for dense layers.
\newline
\newline
The final loss objective to minimize during training of task $t$ is:
\begin{equation}
    \label{final_loss_function}
    L^t = {CL}^t + \alpha \cdot \tau \cdot \sum_{j=1}^L{{DL}^t_j}
    % use lambda instead of alpha
\end{equation}

where $\alpha$ is a tunable hyperparameter and $\tau$ is the temperature parameter:
\[
    \tau = \sqrt{\frac{\sum_{i=0}^t{c^i}}{c^t}} = \sqrt{\frac{{r^t}}{r^t - r^{t-1}}}
\]
as in \cite{LUCIR}. For earlier tasks, $\tau$ is set lower, while for later incremental tasks, $\tau$ is increased. This adjustment is based on the observation that the numerator in the definition of $\tau$ involves a larger number of old classes plus the number of classes in the current task. By increasing the temperature in later stages of incremental learning, the weighting of the distillation loss becomes more significant. This approach addresses the challenge of higher forgetting rates due to the growing number of old classes, ensuring that the model retains knowledge from previous tasks more effectively as new tasks are introduced.
\newline
\newline
We use proxy-based local similarity classifier loss \cite{PODNET} as our classification loss ${CL}^t$. 
\begin{comment}
    The NCA loss is defined as follows:
\[
        \mathcal{L}_{NCA} = -\log \left[ {\frac{\exp{\eta \hat{y}_c}}{\sum_{i \neq c}{\exp{\eta \hat{y}_i}}}} \right]
\]
\end{comment}

It is to be noted that ${DL}^t_j$ is computed using only exemplars. Since exemplars are small in number in comparison to the number of samples in the current task $t$, as suggested in \cite{LUCIR, AFC}, the samples of task $t$ can also be incorporated in the computation of ${DL}^t_j$. However, class-wise significances can be computed only for the classes from the previous tasks and hence are not available for the samples of task $t$ as they belong to a new set of classes. To overcome this limitation and incorporate the samples of task $t$ in the computation of ${DL}_j^t$, we fix the significances of the new classes to be $1$.
\\ \\
\noindent Algorithm \ref{alg:training_algorithm} enumerates the steps to train the model at task $t$.

\begin{algorithm}
    \caption{EXACFS Estimation at Task $t$}
    \label{alg:EXACFS}
    \begin{algorithmic}[1]
    \State \textbf{Input:} $E^{0 \sim t-1}$: exemplars for task $t$
    \State $D^t$: current samples for task $t$
    \State $\widetilde{S}_j^{t-1, c}$: normalized class-wise feature significances at task $t-1$, $j \in \{1, \ldots, L\}$
    \State $\beta \in [0,1]$
    \State \textbf{Output: } $\widetilde{S}_j^{t, c}$ : normalized class-wise feature significances at task $t$, $j \in \{1, \ldots, L\}$
    \State \textbf{Initialization: } $S_j^{t,c} \xleftarrow{} 0$, $j \in \{1, \ldots, L\}$
    \State \For{(k, (x, y)) \textbf{in} $\text{ENUMERATE}(D^t \bigcup E^{0 \sim t-1}$)} \do
    {
        \State \quad $\hat{y} = g^t(P^t(F^t(x)))$
        \State \quad Compute $CL^t(\hat{y}, y)$
        \State \quad Perform backpropagation and compute gradients
        \State \quad \For{$j = 1 \sim L$} \do
        {
            \Statex \quad \quad $S_j^{t,y} \xleftarrow{} S_j^{t,y} + || \nabla_{k}^{y} f_j^t ||^2$
        }
    }
    \State $S_j^{t,c} \xleftarrow{} \frac{S_j^{t,c}}{\sum_{k=1}^{r^t}S_j^{t,c}}$, for $c = 1 \sim {r^t}$
    \State $\widetilde{S}_j^{t,c} = \beta \cdot \widetilde{S}_j^{t-1, c} \, + \, (1 - \beta) \cdot S_j^{t, c}$          
    \end{algorithmic} 
\end{algorithm}

\begin{algorithm}
  \caption{Training Algorithm}
  \label{alg:training_algorithm}
  \begin{algorithmic}[1]  
    \State \textbf{Input:} $E^{0 \sim t-1}$: exemplars from previous tasks
    \State $D^t$: new data for the current task 
    \State $\widetilde{S}_j^{t-1, c}$: per-class feature significances computed at task $t-1$, 
    \State $M^{t-1}$: model obtained at the end of task $t-1$
    \State{\bf Output:} Trained model $M^{t}$ and the per-class feature significances $\widetilde{S}_j^{t, c}$
    \State \textbf{Initialization:} $M^t \xleftarrow{} M^{t-1}$
    \State \textbf{Repeat} 
    {
    \Statex \quad \For{$(x,y) \in D^t \bigcup E^{0 \sim t-1}$}
    {
            \Statex \quad \quad Obtain and store $f_j^t$'s and $\hat{y}$ by forward propagating $x$ through $M^t$
            \Statex \quad \quad Obtain and store $f_j^{t-1}$ by forward propagating $x$ through $M^{t-1}$
            \Statex \quad \quad Compute $\Delta{f_j^t}$ (see equation \eqref{eq:distillation_loss_term})
             \Statex \quad \quad Calculate the distillation losses $DL_j^t$
            \Statex \quad \quad Calculate the classification loss ${CL}^t(\hat{y}, y)$
            \Statex \quad \quad Obtain the total loss using equation \eqref{final_loss_function} 
            \Statex \quad \quad Backpropagate and update model params
    }
    \State \textbf{Until} Convergence
    }
        \State Compute $\widetilde{S}_j^{t, c}$ using Algorithm \ref{alg : EXACFS}
  \end{algorithmic}
\end{algorithm}

\section{Experiments}
\label{sec:Experiments}
In this section, we demonstrate the effectiveness of our approach EXACFS by benchmarking it against SOTA methods on standard datasets. Various ablation studies are also presented.
\subsection{Datasets}
To quantitatively assess the performance of our algorithm, we used the following benchmark datasets: CIFAR-100 and ImageNet100. The CIFAR-100 dataset \cite{CIFAR100} comprises 60,000 color images across 100 classes, each with a resolution of $32 \times 32$. It is split into 50,000 training images and 10,000 testing images. ImageNet100 \cite{ICARL} is a subset of the original ImageNet \cite{IMAGENET} dataset, which includes over a million images spanning 1,000 classes. In ImageNet100, 100 classes are randomly selected, with each class containing approximately 1,300 color images. The sampling of ImageNet100 is performed as described in \cite{LUCIR, PODNET}.

\subsection{Experimental Settings}
 For all datasets, we used the SGD optimizer with a momentum of 0.9 and an initial learning rate of 0.1, along with a cosine annealing schedule and a learning rate decay of 0.1. We trained a ResNet-32 for 160 epochs with an L2-regularization weight of $5 \times 10^{-4}$ for CIFAR-100, and a ResNet-18 for 90 epochs with an L2-regularization weight of $1 \times 10^{-4}$ for ImageNet100, followed by an additional 20 epochs of class-balanced fine-tuning \cite{LUCIR} for both datasets.
\newline
\newline
Following the heuristics proposed in \cite{PODNET}, we normalized each feature map using its Frobenius norm to compute the distillation loss. This loss is computed on features from all the four convolutional stages of the ResNet architectures, excluding the dense layer features. We set the alpha value in equation \eqref{final_loss_function} to 4.0 for CIFAR-100 and 10.0 for ImageNet100. For constructing the exemplar memory, we used herding-based selection \cite{ICARL}, with a fixed memory of 20 exemplars per class and $\beta$ set to 0.4. All models are trained on an NVIDIA GeForce RTX 4090.

\subsection{Evaluation Protocol}
We adhere to the evaluation protocol outlined in \cite{ICARL} and report the Average Incremental Accuracy, which measures the average accuracy of the model across all incremental stages. The evaluation begins with the 'Base Training' task, where the model is initially trained on half of the total classes. Subsequently, we incrementally train the model across $N$ tasks, where $N$ is determined by the number of classes added per task. For both CIFAR-100 and ImageNet100, the increment sizes are set to 1, 2, 5, and 10. We used three different random but fixed class orderings for CIFAR-100 and one random but fixed ordering for ImageNet100. Except for the initial task, which includes half of the dataset’s total number of classes, the remaining classes are evenly distributed across subsequent tasks. We evaluate the model on the test data from all classes encountered up to the current task. For CIFAR-100, the Average Incremental Accuracy is reported as the $\text{mean} \pm \text{std}$ across three independent orderings.
\subsection{Comparative Results}
Since our proposed EXACFS is a distillation-based approach, we benchmark its performance against recent SOTA distillation methods for a fair comparison. We also include a few replay-based methods, such as BiC \cite{BIC}, Mnemonics \cite{Mnemonics}, as well as GDumb \cite{GDumb}, a popular gradient-based replay method. Table \ref{tab:Cifar} presents the results of EXACFS alongside these methods on CIFAR-100. Our method consistently achieves the highest accuracy across all incremental settings and demonstrates superior performance even as the number of tasks increases. EXACFS shows minimal performance degradation with the increasing number of tasks, highlighting its robustness under challenging conditions. Only GDumb \cite{GDumb} exhibits similar resilience, although its performance is approximately $1.5\%$ lower than that of our method for the smallest increment of $1$ class per task that corresponds to the largest number of $50$ tasks. GeoDL is the second best performing method after EXACFS when the number of incremental tasks are lesser.
\newline
\newline
For the ImageNet100 dataset, we primarily compare EXACFS against distillation-based approaches. Table \ref{tab:Imagenet100} shows the performance of EXACFS relative to other methods on ImageNet100. EXACFS performs significantly better in scenarios involving a higher number of incremental tasks, where models are generally more susceptible to catastrophic forgetting. In fact, it shows over $10\%$ gain in comparison to the next best performing method PODNet. It also demonstrates a slight edge over other SOTA methods in scenarios with fewer incremental tasks. Notably, EXACFS surpasses the nearest competitor by $2.06\%$, $1.41\%$, and $2.06\%$ for $25$, $10$, and $5$ incremental tasks, respectively.

\subsection{Ablation Studies}
We consider the following 4 ablation studies viz. (i) impact of per-class feature significance, (ii) effect of restricting distillation, (iii) best exemplar sampling strategy and (iv) impact of memory budget. All the ablation studies have been carried out on CIFAR-100 dataset.

\noindent \subsubsection{Per-Class Feature Significance} 

\begin{figure}[H]
    \centering
    \includegraphics[width=0.9\linewidth]{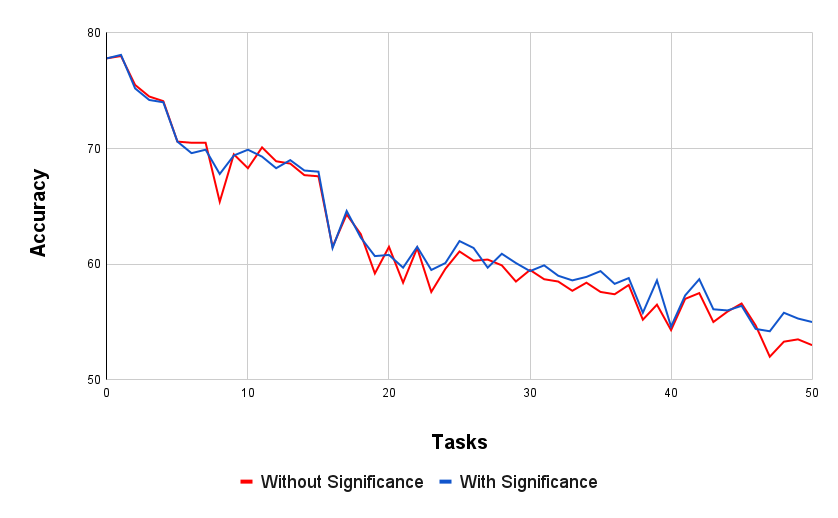}
    \caption{\textbf{Test accuracy of $10$ classes from the set of classes used for base training across the incremental tasks.}}
    \label{fig:best_acc_bc}
\end{figure}

\noindent In this study, we underscore the crucial role of the per-class significance factor $\widetilde{S}_j^{t, c}$ in enhancing the distillation loss in EXACFS to improve model stability. We evaluate and plot the test accuracies of a set of 10 classes from the oldest task (task 0) under a highly susceptible scenario for catastrophic forgetting, where each task contains only $1$ class. Figure \ref{fig:best_acc_bc}
presents the results, revealing that as these 10 classes age over numerous incremental tasks, the model incorporating per-class significance for distillation significantly outperforms the model without it by a margin of $2\%$. In this context, "model without per-class significance" refers to treating all classes equally, implying that $\Delta f_j$ in equation \eqref{eq:distillation_loss_term} is uniformly weighted across all exemplars from different classes.

\begin{comment}
    \begin{figure}
    \centering
    \includegraphics[width=0.9\linewidth]{images/average_acc_base_classes.png}
    \caption{Average test accuracy of classes used for base training across the incremental tasks.}
    \label{fig:avg_acc_bc}
    \end{figure}
\end{comment}

\noindent \subsubsection{Restricting Distillation}

\noindent As previously discussed, our method, EXACFS, is designed to be applicable at every level of feature extraction within the model. In contrast, models such as LUCIR \cite{LUCIR} apply distillation solely at the final feature representation stage. In this study, we compare these two approaches within our architecture and demonstrate that applying distillation across all layers results in less forgetting. Table \ref{tab:distil_stage4} indicates that incorporating distillation across all stages of the model yields a significant improvement of $1.2\%$ on average and around $3\%$ specific to the crucial setting of $50$ incremental tasks in performance compared to applying distillation only at the final stage. It is important to note that the ResNet architecture used in this study consists of four stages of Residual Blocks.

\noindent \subsubsection{Exemplar Selection} 

\noindent We have employed the widely-used herding-based selection method \cite{ICARL} for selecting exemplars. However, SOTA methods, such as \cite{SSIL}, utilize random selection for this purpose. In this study, we aim to determine the optimal exemplar selection strategy for our proposed method by evaluating several approaches. The strategies considered include herding-based selection, random selection, and closest-to-mean selection.

\noindent In the herding-based selection, at iteration $k$, the $k^{th}$ exemplar of the class is chosen such that the mean of the selected $k$ exemplars is closest to the class prototype. Random selection uniformly selects exemplars from a class. In the closest-to-mean selection, the exemplars closest to the class prototype are selected. Table \ref{tab:exemplar_selection} presents the performance of EXACFS across these various exemplar selection strategies under the setting of $5$ incremental tasks. The results indicate that all the examined exemplar selection strategies perform comparably well.

\noindent \subsubsection{Memory budget} 

\begin{figure}[H]
    \centering
    \includegraphics[width=0.9\linewidth]{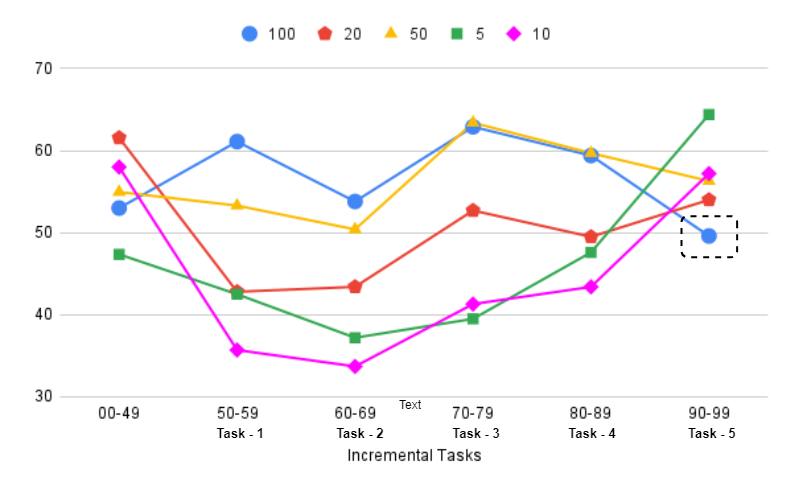}
    \caption{\textbf{Per-task test accuracies of the model trained on the last incremental task across different memory budgets}}
    \label{fig:stability_plot}
\end{figure}

\noindent By memory budget, we refer to the number of exemplars stored for each class. To evaluate the impact of memory budget on EXACFS, we conducted experiments using memory budgets of $5$, $10$, $20$, $50$ and $100$ exemplars per class under the setting of $5$ incremental tasks. Table \ref{tab:memory_budget} presents the effect of varying memory budgets on EXACFS. As the number of exemplars increases, performance improves up to a threshold of $20$ exemplars, after which it drops significantly. This decline can be attributed to the following reason: distillation loss is imposed to address the imbalance between the number of exemplars and the samples from the current task. When the number of exemplars increases, thereby reducing this imbalance, continuing to impose feature similarity through distillation naturally inhibits the model's plasticity. Reduced plasticity subsequently lowers performance, as clearly reflected in the table for cases with $50$ and $100$ exemplars. This reasoning is also supported by Figure \ref{fig:stability_plot}, wherein the per-task test accuracies of the model trained on the last incremental task is plotted. It can be observed that the model's performance on the most recent task comprising of classes $90$-$99$ in the case of $100$ exemplars (the scatter point enclosed by a dotted box in the figure) has drastically come down, degrading the plasticity.

\section{Conclusions}
In this work, we introduced EXACFS, a novel method for training neural networks in class incremental settings. The key innovation of our approach is the introduction of class-wise feature significance, which determines the importance of each feature in correctly classifying input samples of a particular class. We use exponential averaging to incorporate the history of per-class feature significance values. Our experiments on benchmark datasets such as CIFAR-100 and ImageNet100 demonstrate the effectiveness of our approach.
\newline 
\newline
A limitation of this work is the need to store two models, the previous and the current, to activate feature distillation. Many SOTA feature distillation methods also require storing multiple models. While feature distillation is a promising direction for mitigating catastrophic forgetting, the requirement of storing two or more models can limit its deployment in systems with restricted memory budgets. Future work could explore whether storing and using class prototypes of exemplars could replace the need for storing the entire previous model.

% --------------CIFAR-100---------------
\begin{table*}
\caption{\textbf{Comparative Analysis on CIFAR-100 dataset}}
\label{tab:Cifar}
\centering
\scalebox{1.0}{
\begin{tabular}{@{}lcccc@{}}
\toprule
 & \multicolumn{4}{c}{CIFAR100} \\
  & 50 tasks & 25 tasks & 10 tasks & 5 tasks \\
 \multicolumn{1}{r}{New classes per task} & 1 & 2 & 5 & 10 \\
 \midrule
iCaRL~\cite{ICARL} & 44.20$\pm$0.98 & 50.60$\pm$1.06 & 53.78$\pm$1.16 & 58.08$\pm$0.59 \\ 
BiC~\cite{BIC} & 47.09$\pm$1.48 & 48.96$\pm$1.03 & 53.21$\pm$1.01 & 56.86$\pm$0.46 \\
% LUCIR\,{\scriptsize (NME)}~\cite{hou2019lucir} & 48.57$\pm$0.37 & 56.82$\pm$0.19 & 60.83$\pm$0.70 & 63.63$\pm$0.87 \\
LUCIR\,~\cite{LUCIR} & 49.30$\pm$0.32 & 57.57$\pm$0.23 & 61.22$\pm$0.69 & 64.01$\pm$0.91 \\
Mnemonics~\cite{Mnemonics} & - & 60.96$\pm$0.72 & 62.28$\pm$0.43 & 63.34$\pm$0.62 \\
GDumb~\cite{GDumb} & 59.76$\pm$1.49 & 59.97$\pm$1.51 & 60.24$\pm$1.42 & 60.70$\pm$1.53 \\
% PODNet\,{\scriptsize (NME)}~\cite{douillard2020podnet} & 56.78$\pm$0.41 & 59.54$\pm$0.66 & 63.27$\pm$0.69 & 65.32$\pm$0.65 \\
PODNet\,~\cite{PODNET} & 57.86$\pm$0.38 & 60.51$\pm$0.62 & 62.78$\pm$0.78 & 64.62$\pm$0.65 \\
GeoDL~\cite{GeoDL} & 52.28$\pm$3.91 & 60.21$\pm$0.46 & 63.61$\pm$0.81 & 65.34$\pm$1.05 \\ 
SSIL~\cite{SSIL} & 53.64$\pm$0.77 & 58.02$\pm$0.79 & 61.52$\pm$0.44 & 63.02$\pm$0.59 \\
\hdashline
EXACFS (\textbf{OURS}) & \textbf{61.1$\pm$0.75} & \textbf{62.75$\pm$0.8} & \textbf{64.05$\pm$1.05} & \textbf{65.48$\pm$1.08} \\
\bottomrule
\end{tabular}
}
\end{table*}

% --------------------Table for ImageNet100-----------------------
\begin{table*}[h]
\caption{\textbf{Comparative Analysis on ImageNet100 dataset}}
\label{tab:Imagenet100}
%\vspace{0.1in}
\centering
\scalebox{1.0}{
\begin{tabular}{c c c c c}
 \hline
 & \multicolumn{4}{c }{ImageNet100} \\
 % \cline{0-1}
 & 50 tasks & 25 tasks & 10 tasks & 5 tasks \\
 % \hline
 New classes per task & 1 & 2 & 5 & 10 \\
 \hline
 iCaRL~\cite{ICARL}         & 54.97 & 54.56 & 60.90  & 65.56 \\
 % \hline
 BiC~\cite{BIC} & 46.49 & 59.65 & 65.14  & 68.97 \\
 % \hline
 LUCIR\,~\cite{LUCIR}     & 57.25 & 62.94 & 70.71  & 71.04 \\
 % \hline
 Mnemonics~\cite{Mnemonics} & - & 69.74 & 71.37 & 72.58 \\
 % \hline
 PODNet\,~\cite{PODNET}                & 62.48 & 68.31 & 74.33 & 75.54 \\
 % \hline
 GeoDL~\cite{GeoDL} & - & 71.72 & 73.55 & 73.87 \\
 % AFC~\cite{AFC} & 72.08 & 73.34 & \textbf{\textcolor{red}{75.75}} & \textbf{\textcolor{red}{76.87}} \\
 \hdashline 
 EXACFS (\textbf{OURS})\,                & \textbf{72.50} & \textbf{73.78} & \textbf{74.96} & \textbf{75.93} \\
 \hline
\end{tabular}
}
%\vspace{0.2cm}
\end{table*}

% figure: best classes accuracy

% memory budget table
\begin{comment}
    \begin{table}[H]
        \caption{Impact of Memory Budget}
        \centering
        \begin{tabular}{|c| c c c c c|}
        \hline
        \textbf{Memory Budget}                & 5 & 10 & 20 & 50 & 100 \\
        \hline
        \textbf{Average Incremental Accuracy} & 56.84 & 63.17 & 65.48 & 65.01 & 63.95 \\
        \hline
        \end{tabular}
        \label{tab:memory_budget}
\end{table}    
\end{comment}

\begin{table}[H]
        \caption{\textbf{Impact of Memory Budget}}
        \centering
        \begin{tabular}{|c | c|}
        \hline
        \textbf{Memory Budget} & \textbf{Average Incremental Accuracy} \\
        \hline
         5                      & 56.84 \\
        \hline
         10                     & 63.17 \\
         \hline
         20                     & 65.48 \\
         \hline
         50                     & 65.01 \\
         \hline
         100                    & 63.95 \\
        \hline
        \end{tabular}
        \label{tab:memory_budget}
\end{table}

% memory budget line plot showing stability is maintained

% exemplar selection table
\begin{table}[H]
    \caption{\textbf{Exemplar Sampling}}
    \centering
    \resizebox{0.75\columnwidth}{!}{ % Resize to column width
        \begin{tabular}{|c|c|c|c|}
        \hline
        \textbf{Exemplar selection} & Herding & Random & Closest-to-mean \\
        \hline
        \textbf{Average Incremental Accuracy} & 65.48 & 65.44 & 65.47 \\
        \hline
        \end{tabular}
    }
    \label{tab:exemplar_selection}
\end{table}

% distillation stage table
\begin{table}[H]
    \caption{\textbf{Restricted Distillation vs Distillation Throughout}}
    \centering
    \begin{tabular}{|c| c c c c|}
    \hline
    \textbf{New classes per task}        & 1 & 2 & 5 & 10 \\
    \hline
    \textbf{Distillation only at last stage} & 58.93 & 62.33 & 63.75 & 64.98 \\
    \hline
    \textbf{Distillation at all stages} & 61.85 & 63.55 & 65.10 & 66.56 \\
    \hline
    \end{tabular}
    \label{tab:distil_stage4}
\end{table}

\section*{Acknowledgements}
This work is dedicated to Bhagawan Sri Sathya Sai Baba, the Founder Chancellor of Sri Sathya Sai Institute of Higher Learning.

\bibliographystyle{unsrt}  

%%% Comment out this section when you 
\bibliography{references}
\end{document}